# Pedestrian Attribute Recognition in Video Surveillance Scenarios Based on View-attribute Attention Localization

Weichen Chen[1]    Xinyi Yu[1]    Linlin Ou[1]

[1] College of Information Engineering, Zhejiang University of Technology, Hangzhou 310023, China

**Abstract:**    Pedestrian attribute recognition in surveillance scenarios is still a challenging task due to the inaccurate localization of specific attributes. In this paper, we propose a novel view-attribute localization method based on attention (VALA), which utilizes view information to guide the recognition process to focus on specific attributes and attention mechanism to localize specific attribute-corresponding areas. Concretely, view information is leveraged by the view prediction branch to generate four view weights that represent the confidences for attributes from different views. View weights are then delivered back to compose specific view-attributes, which will participate and supervise deep feature extraction. In order to explore the spatial location of a view-attribute, regional attention is introduced to aggregate spatial information and encode inter-channel dependencies of the view feature. Subsequently, a fine attentive attribute-specific region is localized, and regional weights for the view-attribute from different spatial locations are gained by the regional attention. The final view-attribute recognition outcome is obtained by combining the view weights with the regional weights. Experiments on three wide datasets (RAP, RAPv2, and PA-100K) demonstrate the effectiveness of our approach compared with state-of-the-art methods.

## 1   Introduction

With the expansion of surveillance technology, video surveillance systems have been widely used in many domains, e.g., security, criminal investigation, and traffic. Therefore, pedestrian attribute recognition in video surveillance has great potential, facilitating the evolution of person retrieval and person re-identification. Pedestrian attribute recognition aims to make predictions for a group of attributes, e.g., gender, age, wearing a dress. Recently, methods[1, 2] based on convolutional neural networks (CNN) have achieved great success in recognizing pedestrian attributes.

However, some difficulties and challenges in pedestrian attribute recognition still exists, such as multi-view change, low resolution, low illumination, and occlusion in complex backgrounds. A solution is to guide the recognizing process via prior knowledge. Previous works[2-4] attempt to take body parts or pose information as prior knowledge. Bourdev et al.[3] proposed a body-part detector to assist attributes recognition units. Similarly, Li et al.[2] employed human posture as the prior knowledge to recognize attributes of body parts. However, these methods highly depend on external human body detectors, which bring in extra inferring time and computational cost.

Since images are recorded by monitoring cameras, multi-view change becomes the most severe challenge under dynamic scenarios. Several associated problems can be caused by viewpoint change, such as, region change problem for the same attribute of the identical pedestrian from different views. As depicted in Fig. 1, for the same pedestrian identity, the spatial areas of the hat attribute (group(b)) are larger from the side views (3rd row and 4th row) than that from the front view (1st row). On the contrary, the shoulder bag attribute in group(a) occupies a larger spatial area from the front view. Thus, there is a relationship between a specific pedestrian attribute and different image views. We can utilize this relationship to guide the attribute recognition process and alleviate the viewpoint change problem. Therefore, besides taking body parts as prior knowledge, view information can be considered another efficient clue to guide the training process to focus on a specific attribute. Furthermore, since the importance of each view is different for a specific attribute, we can utilize view information by giving a big weight to the most important view and a small weight to the less important view. Moreover, the weights are predicted by the view prediction branch in our paper.

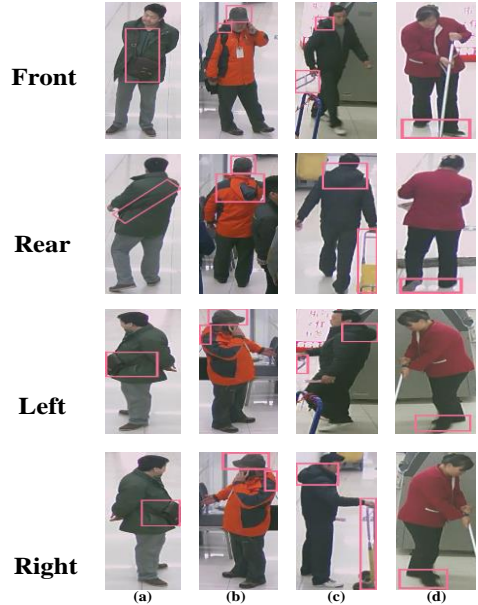

Fig. 1    Images of the same pedestrian identity from different views. (a-d) Example images are from the RAPv2 dataset[5], and typical attributes including *shoulder bag*, *shoes*, and *hat* are marked by pink bounding boxes.

When the same pedestrian attribute of images is under different views, corresponding regions of the attribute are different. Therefore, though the training process can concentrate on recognizing a specific attribute through leveraging view information, it is still necessary to obtain the attribute spatial location information for better recognizing performance. Attention mechanism[6-8] is introduced to localize attribute-related regions and give the weights for different spatial locations in recent methods. These methods[7, 8] usually yield the attention masks from feature layers and then multiply attention masks with the corresponding feature maps to get the positional weights of the attributes. Giving a region a bigger weight denotes that the region is most likely to include this attribute, and weights can be visualized like Fig. 2. As shown in Fig. 2(b), the learned attention mask attends a broad region and points out shirt, backpack, and jeans attributes. However, the attention mask is inaccurate for localizing a specific attribute and mixes the regions of different attributes to deal with spatial information and channel inter-dependencies insufficiently. Later studies[9, 10] demonstrate the decisive capabilities of channel attention and spatial information to generate attentive weighted regions. Concretely speaking, SE-Net[9] received cross-channel relationships in feature maps by learning modulating weights per-channel. This was followed by CBAM[10] which accomplished channel attention and spatial attention in two processes to enrich attentive regions. As illustrated in Fig. 2(c,d), when recognizing *jeans*, attentive regions produced by SE-Net[9] and CBAM[10] focus on specific positions. Nevertheless, SE-Net[9] only considers the importance of channel attention, and both of them[9, 10] reduce the channel dimension. Hence, these two modules are more likely to cause information loss, and the generated attention areas have deviations to some extent. To avoid feature loss and preserve spatial attribute information, our attention branch considers fusing the two processes from CBAM[10] into one process to localize and generate attentive weights for different regions of specific attributes from different views.

A novel method is proposed in this paper, namely View-attribute Localization based on Attention (VALA), which exports view prediction and attribute inference of pedestrian images recorded by surveillance cameras. By making full use of the importance of different views for each specific attribute, the view information is adopted as prior knowledge to guide the training process to focus on specific attributes. The view prediction branch is designed to utilize view information by predicting view weights as the confidences for attributes from different views. Additionally, view information is holistic, so the view predictor is placed in shallow layers, and predicts four view weights (incl. front, rear, left, right) for low-level attribute features. Subsequently, the view weights are combined with shallow features to form specific view-attributes. The joint process of views and attribute features makes it easy to identify the attributes associated with one of the views when the images are in this view. Since the spatial locations of a specific attribute also change with the change of views, regional attention is proposed to acquire attribute location information and localize precise attribute-related regions. The regional attention branch restricts the pedestrian position into a frame first, and then localizes view-attribute regions to output regional weights. Specifically, the regional attention is divided into three small branches: *height* branch, *width* branch, and *ratio-balance* branch. The first two branches are responsible for embedding spatial attribute information of feature maps along height dimension and width dimension, and channel inter-dependencies are captured after aggregating spatial information from two spatial dimensions. Then two attention maps are generated as regional weights. After balancing the ratio of the two attention maps by the *ratio-balance* branch, a fine attribute-related region and regional weights are generated. Finally, the evaluating result is obtained by multiplying the view weights by the regional weights.

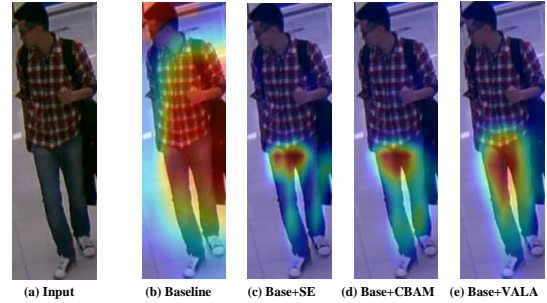

Fig. 2 Visualization of attribute attentive areas produced by different attention modules. The example image is from the RAPv2 dataset[5], and (b-d) visual attention-related regions of the jeans attribute are generated by a single baseline (refer to ResNet50[11] here), SE module[9], CBAM module[10] and our VALA module respectively.

The contributions of this paper are presented as follows:

- We establish an end-to-end trainable framework that takes view information as prior knowledge to compose specific view-attributes in deep network supervision.

- We propose regional attention to further localize specific view-attributes rather than classify attribute regions from different views by manual operation.

- We conduct extensive experiments on three public pedestrian attribute recognition datasets, i.e, RAP[12], RAPv2[5], and PA-100K[13], which clearly illustrate the competitive generalization ability and potentiality of our method.

## 2 Related Work

### 2.1 Pedestrian Attribute Recognition

Pedestrian attribute recognition can be seen as a kind of mid-level task of pedestrian analysis in video surveillance, which may provide important information for high-level person related tasks, such as person re-identification[14-16], pedestrian detection[17], person tracking[18], person retrieval

[19, 20], video action recognition[21-24], and action segmentation[25-28].

Early pedestrian attribute recognition methods such as HOG[29], SVM[30] focus on hand-crafted features, but the performance produced by these conventional methods is far from satisfactory. With the rapid development of convolution neural networks, recent methods based on CNNs have achieved great success.

These methods[1, 2, 4, 6, 8, 31-35] can be classified into five categories:

**Global-based:** ACN[1] used a CNN model to jointly learn attributes and calculated the loss for each attribute. DeepMAR[31] regarded the attribute learning process as a multi-label task and computed the loss of all attributes via sigmoid cross-entropy loss. However, global-based methods are out of application due to a lack of consideration of local fine-grained features.

**Part-based:** PGDM[2] adopted a pose estimation model to obtain information about auxiliary human body-parts. LG-Net[4] applied EdgeBoxes[36] to generate region-proposals for local attribute features. DTM+AWK[32] leveraged pose key points as auxiliary information to help the main module locate proper attribute regions. These part-based methods that capture local body parts and posture features improve the performance significantly, but bring extra inferring time and computational cost from external part localization modules.

**Relation-based:** JRL[33] exploited the inter-dependencies among attributes to conduct joint recurrent learning using a CNN-RNN model. GRL[34] explored the intra-group and inter-group relationships of attributes and then divided all attributes into several groups to recognize. These methods improve performance by extracting the relationship between attributes, but the modules of these methods are usually complicated, and the parameters are hard to control.

**Attention-based:** DIAA[6] introduced multi-scale attention to deal with the problem of attribute imbalance. Da-HAR[8] carried a coarse-to-fine attention mechanism to reduce irrelevant areas and improve the discriminative power for attribute recognition. As mentioned in Section 1, attention-based methods are always influenced by complex backgrounds and surroundings, making attention masks fail to obtain the position of a specific attribute.

**Attribute-based:** ALM[35] induced a localizing method for specific attributes to discover the most discriminative attribute regions.

Later studies are dedicated to solving the problems in pedestrian attribute recognition, but it is still likely to have many aspects of improving the recognition performance in the future. Presently, the attention of investigators gradually diverts from localizing generic attributes to specific attributes. Our method proposes regional attention to localize composed specific view-attributes and obtain more precise attribute regions to achieve a greater attribute recognition effect.

### 2.2 Multi-view Information

Besides occlusion and blurring, viewpoint change problem is difficult to handle under dynamic scenarios. Since the importance of different views to each specific attribute is different, it seems to have a relationship between attributes and views. Therefore, by making full use of the relationship, visual clues are available to be used as auxiliary supervision to help with attribute recognition. Some methods[37-39] already attempt to leverage viewpoint information in different fields. As views are global, partial view-specific information may be ignored during feature extraction stages. Feng et al. [37]offered a view-specific deep network, which extracted view clues more comprehensively to verify the great performance of the view-feature model in the re-identification field. Multi-view information is also considered in the face verification domain. Farfade et al. [38]proposed a deep and dense face detector to detect faces in a wide range of views and discovered that different views could help face detection to handle the occlusion problem and capture more facial features to some extent. In sentiment analysis research, Sadr et al. [39]decided to aggregate features extracted from heterogeneous neural networks by using multi-view classifiers, which enhanced the overall performance of document-level sentiment analysis. Our method takes view information as prior knowledge in the pedestrian attribute recognition field. View information is leveraged in our paper by a view prediction branch, which can guide the deep feature extraction process to focus on specific attributes.

### 2.3 Attention Localization

When observing an object in daily life, people are always involved in its relevant resource, but ignore the irrelevant areas. Originating from human visual behaviors, an attention mechanism is introduced to pay more attention to related regions. Zhu et al. [40] first applied the attention mechanism to the pedestrian attribute recognition field and revealed that the core of the attention mechanism is to find the most representative attribute-related regions by giving the largest weight[41]. Liu et al. [13]carried an attention module to fuse multi-scale features from multiple levels to yield attention maps. Thus, attention maps from higher blocks can cover more extensive attribute regions, and lower blocks can concentrate on smaller attribute regions of the input images. However, complex surroundings and backgrounds always prevent usual attention methods localizing related regions accurately. To alleviate the challenge, Yaghoubi et al. [7] introduced a coarse attentive body segmentation module, which multiplies features and attention masks to discriminate between the foreground and the background. Whereas, their attention masks fail to take into consideration the attribute-specific context, so they[7] find that coarse attention is not as helpful as their expectation. When the attention mechanism needs to localize specific attributes, a helpful solution is to embed inter-channel and spatial information of specific attribute feature maps. SE-Net[9] squeezed each 2D feature map to efficiently build inter-dependencies among channels. CBAM[10] further advanced this idea by encoding spatial information and handling channel extraction and spatial encoding in two independent processes. However, SE-Net only considers the importance of channel attention, and CBAM does not consider the correlation between the channel and spatial information. These two methods are apt to cause information loss. Since the related regions of a specific attribute are different in each view, we apply an attention mechanism to localize more precise regions. Taking inspiration from CBAM, two processes are aggregated into one in our regional attention branch, which captures channel inter-dependencies and preserves spatial information with the help of global max pooling and global average pooling.

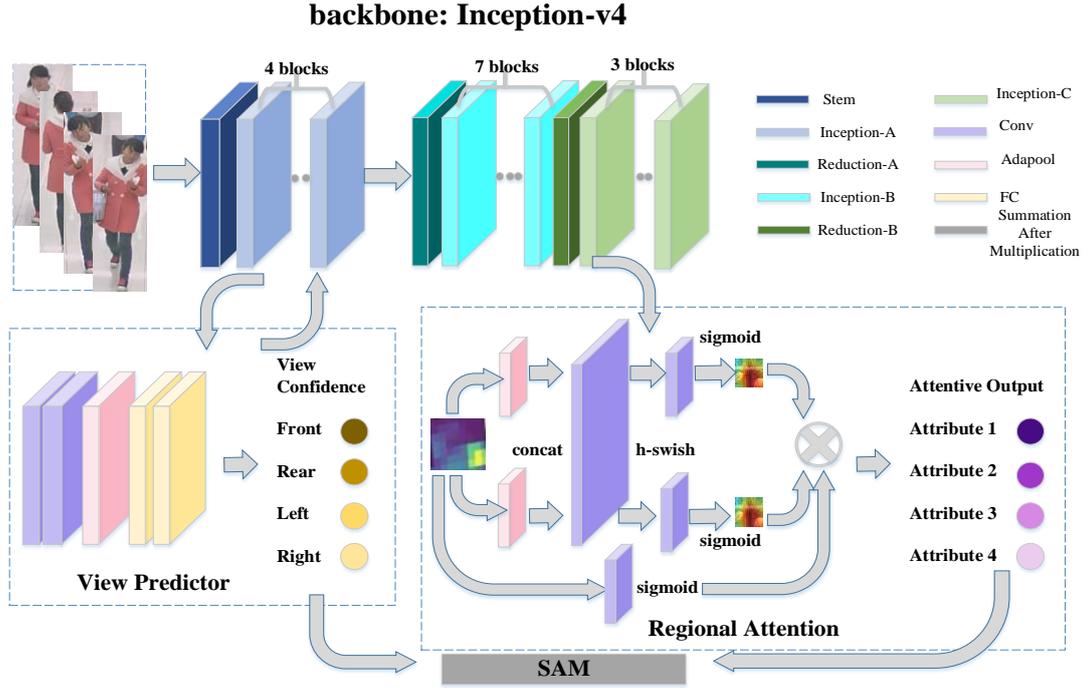

Fig. 3　The overall framework of the VALA method. The primary network follows the structure of Inception-v4[42], and Inception-ResNet-v2[42] network can be considered to replace Inception-v4 in training for rapid training speed. View prediction branch is constructed after Inception-A to gain four view weights, while the regional attention is built after Inception-C to localize attribute-corresponding regions. Eventually, the result is obtained by the above two branches via multiplication and summation. Different modules are annotated in different colors in the framework.

## 3　Proposed Method

The overall framework of VALA is illustrated in Fig. 3. VALA consists of the main network, a view prediction branch, and a regional attention branch. The main network is built to accomplish feature extraction first. Then, the view prediction branch utilizes the features from shallow network layers to predict four view weights for specific attributes from different views. View weights are then fed back to shallow layers to compose view-attributes. To obtain the related regions of a specific view-attribute, the attention branch named *regional attention* is introduced to output regional weights for the view-attribute from different spatial locations. Finally, the attribute recognition outcome is gained by multiplying the view weights by with regional weights.

### 3.1　Network Architecture

During the process of pedestrian attribute recognition, global attribute feature extraction relies on large convolutional kernels, while local attribute feature extraction is dependent on small convolutional kernels. Therefore, Inception-v4[42] is adopted as our primary network, which possesses different sizes of convolutional kernels for different scales of features in the same layer to obtain both global and local features simultaneously. Meanwhile, Inception-v4 is sufficiently deep for extracting semantically stronger features and has a more uniform simplified architecture with more inception modules than Inception-v1-3[43-45].

Inception-v4 mainly consists of Stem, Inception-A, Reduction-A, Inception-B, Reduction-B, and Inception-C blocks. Inspired by Google-Net (Inception-v1[43]), which has two auxiliary classifiers, those two auxiliary branches are replaced with the view predictor and the regional attention respectively in the proposed method, but the positions of our two branches do not correspond to the original locations in Google-Net. Furthermore, the average pooling layer, the dropout layer, and the final softmax in the original network are eliminated. A BN layer is used to normalize the final attribute recognition units directly instead.

Since the Inception-v4 network tends to be very deep, it is natural to combine inception architecture with residual connections, which can accelerate the training process of the inception network and explore deeper feature layers simultaneously. Therefore, the Inception-ResNet-v2[42] network can be considered to replace Inception-v4 in training to speed up the inference process of our model.

### 3.2　View Predictor

When monitor cameras are working, the positions of a specific attribute from recorded images vary across different views. Therefore, the relationship between views and attrib-utes exists, which can be utilized to help the recognition process focus on specific attributes. Based on the shared feature maps output by Inception-A blocks, a view prediction branch is designed to utilize the view information in our method, and the architecture of the view prediction branch is shown in Fig. 4. As shown in Fig. 4, for input feature maps $F_1 \in R^{384 \times 17 \times 17}$, pooling and convolution layers are added to remove redundant information firstly. Subsequently, the convolution layer is followed by an adaptive average pooling layer which down-samples the feature maps to the intermediate variable $F' \in R^{128 \times 1 \times 1}$. In our view predictor, we apply the convolution and adaptive average pooling layers to reshape feature maps instead of allowing one convolution to resize directly. This operation can help the feature extraction process explore a larger receptive field. Then, $F'$ is passed through two fully-connected layers to output four view weights (incl. front, rear, left and right) as the confidences for specific attributes from different views. Let $Y_{vp1} = [f, b, l, r] \in R^{4 \times 1 \times 1}$ denote predicted view weights, mathematically, $Y_{vp1}$ can be represented by the following equation:

$$Y_{vp1} = \sigma(W_{fc2} \cdot W_{fc1} \cdot F')  \quad (1)$$

where $\sigma$ is the sigmoid function, and $\cdot$ means the dot product of two matrices. $W_{fc1}$ and $W_{fc2}$ are the weight matrices of the two fully-connected layers. Activated by a sigmoid function, the channel dimension of the view weights is then scaled to C, which is equal to the channel dimension of the input feature maps $F_1$ through a designed view parameter. Based on the theory of neural architecture search[46], the view parameter is designed to expand the ratio of these four view weights without changing the value of weights, so the $Y_{vp1}$ can be fed back to Inception-A blocks to combine the extracted features in Inception-A for composing specific view-attributes. The joint process of attribute features and view weights makes specific attributes associated with their related views, making the view-attributes easier to identify when the input images are in the related view. Moreover, the composed view-attribute participates in further deep feature extraction to construct the relevance of shallow and deep layers and supervise the network to focus on specific attributes.

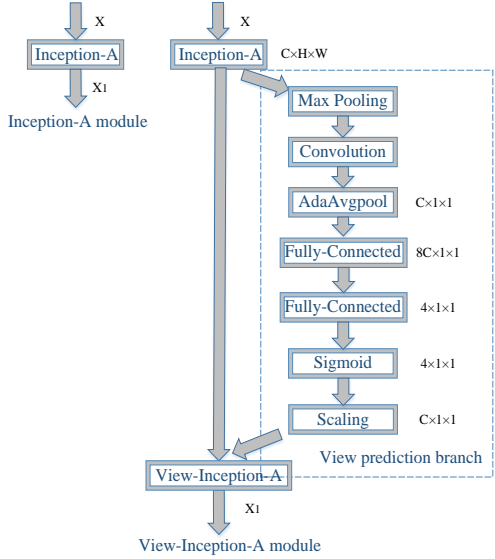

Fig. 4　Module architecture of the Inception-A block with embedding view prediction. Scaling contains the up-sampling operation.

Meanwhile, view weights are also delivered to the final classification units as the contribution of the view prediction branch to guide the attribute recognition process. Replacing the activation function of $Y_{vp1}$ with a softmax function to generate $Y_{vp2}$. Let $Y_{vp2} \in R^{4 \times 1 \times 1}$ denote the delivered view weights, and it can be formulated as:

$$Y_{vp2} = softmax(W_{fc2} \cdot W_{fc1} \cdot F'))  \quad (2)$$

### 3.3　Regional Attention

Since the same attribute lies in different related regions from different views of images, the attention mechanism is applied in our paper to acquire attribute location information and localize precise attribute-related regions. Later studies[9, 10] have confirmed that inter-channel dependencies and spatial information of the input attribute features have an influence on the accuracy of spatial channel integration and attribute-related areas localization. It concretely shows that obtaining the spatial attribute locations benefits the attribute recognition, and adjusting the inter-channel dependencies of different attribute features can handle the attribute imbalance problem. However, previous methods[9, 10] using channel attention and spatial attention do not achieve better performance. For example, CBAM[10] has difficulty handling the relationship between channel attention and spatial attention since these two processes are computed independently. Inspired by the method[10] of constructing spatial attention, we propose an attention mechanism called *regional attention* to utilize global max pooling and global average pooling for preserving spatial information and channel interaction simultaneously.

The structure of our regional attention is shown in Fig. 3. The feature maps output by Inception-C blocks are taken as the input of the regional attention, and the regional attention is factorized into three small branches. Since the pedestrian is shown in a standing posture from the input image, which is not flipped, the whole position of the pedestrian occupies the largest pixels in the height direction. Therefore, before localizing a specific view-attribute, it needs to eliminate background interference and determine the pedestrian position in the height orientation.

Global pooling is often used to encode spatial information, but it squeezes both height and width dimensions in a square shape. However, for the rectangular shapes of pedestrians, using global pooling is prone to causing significant information loss. To get spatial attribute interactions more thoroughly, we apply a spatial extent of adaptive kernels to embed local spatial information along the height dimension. The adaptive kernel is the global max pooling (GMP) which has the capability to extract texture features and eliminate redundant data. Since related regions of different attributes distribute in different pedestrian positions along the width direction, the width branch is built to find a certain region position for a specific attribute. For the width branch, an adaptive global average pooling (GAP) is conducted to capture the spatial information of a specific attribute and localize a coarse related region along the width direction. Concrete localizing operation examples of the height and the width branches are displayed in Fig. 5. The output feature maps from these two small branches can be formulated as follows:

$$GMP(x) = \max_{w}(x(h, i)) \quad (3)$$

$$GAP(x) = \frac{1}{H}\sum_{j=1}^{H}(x(j, w)) \quad (4)$$

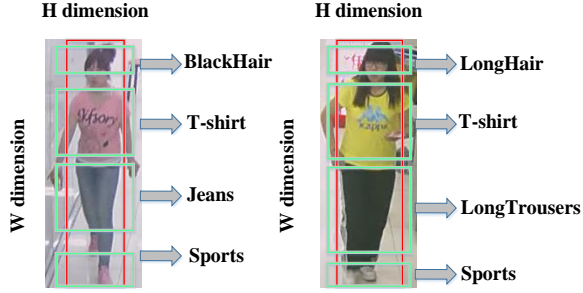

Fig. 5 Examples of localizing operation by height and width branches. Images are from the RAPv2 dataset[5], and red regions are produced in the height dimension, while green regions are generated in the width dimension.

The operation of aggregating spatial information from the above two branches does not reduce any channel dimension and keeps as much spatial information as possible. Unlike the previous shrinking operation[9, 10], retaining the channel dimension can reduce the attribute information loss. Subsequently, we integrate the outputs of the two branches along spatial dimensions to complement spatial attribute information. The integrated result is then passed through a convolution to output the intermediate result $F' \in R^{c \times (h+w) \times 1}$, which is activated by a h-swish function[47]. Based on the ReLU6 function, the h-swish function eliminates latent precision loss and improves the efficiency in the deep network particularly. The intermediate output can be represented by the following equation:

$$F' = h - swish(f(concat(GMP(x), GAP(x))) \quad (5)$$

where $f$ stands for the convolutional operation. Then we split the intermediate vector $F'$ back to two attribute feature vectors $F_1 \in R^{c \times h \times 1}$ and $F_2 \in R^{c \times 1 \times w}$. In order to modulate the inter-channel dependencies for a specific attribute, another two convolution transformations are utilized to augment the channel dimension into $C$, which represents the class number in the final attribute recognition units. Through a sigmoid activation layer, two attention maps are gradually generated as the regional weights of a view-attribute from different spatial regions. However, for localizing the related region of a view-attribute, the required proportion of regional weights in height and width direction is different. Therefore, the third branch, named the *ratio-balance* branch, is constructed to balance the ratio of the regional weights. And all three branches are aggregated together via multiplication. Set $Y_a$ as the adjusted regional weights and $Y_a$ can be written as:

$$Y_a = \sigma(f_3(F)) \times \sigma(f_1(F_1)) \times \sigma(f_2(F_2)) \quad (6)$$

where $\sigma$ is the sigmoid function, $F_1$ and $F_2$ are the split feature vectors from the intermediate vector $F'$ in the height and width dimension, respectively, while $F$ is the initial input attribute features from Inception-C blocks. $f_1$, $f_2$ and $f_3$ represents the convolution operations to handle $F_1$, $F_2$ and $F$.

### 3.4 View-attribute Attention Localization

To utilize view information, we use the view prediction branch to predict view weights to describe the importance of each view for a specific attribute. The number of the view weights $Y_{vp}$ channel is four, and the value of the channel dimension represents the weights occupied by four different views, respectively. Subsequently, the view weights are used in two places. We firstly deliver view weights back to the shallow feature layers to compose view-attributes. Such operation is equivalent to fixing the corresponding view for a specific attribute. Then, the view weights are also fed into the final recognition units to combine with the attribute spatial information.

Moreover, for view-attributes, the related regions are different in different views; we apply the regional attention to capture the spatial location information of view-attributes. The regional attention takes the deep features, which include view features as the input, and this reduces the impact of changing views on positional localization. After inter-channel dependencies encoding and spatial information preservation, the regional attention localizes precise attribute-related regions and outputs regional weights for view-attributes from different spatial locations. Then, the regional weights are fed into the final attribute recognition units as the contribution of regional attention.

The final result is obtained by multiplying view weights of attributes from different views with regional weights of view-attributes from different spatial locations. This can be performed as a weighted sum function that view confidences are used as weights. To multiply each view weight for convenience, we get four portions of regional weights. Next, we let every portion of regional weights $Y_a \in R^{C \times H \times W}$ and one of the view weights $Y_{vp}' \in R^{1 \times 1 \times 1}$ conduct element-wise multiplication, and then sum the whole four multiplied results together. Finally, a BN layer is applied to normalize and estimate the prediction of attributes. Then the attentive location of a specific view-attribute can be realized.

## 4  Experiments

We assess our method on three public datasets, including RAP[12], RAPv2[2], and PA-100K[13]. Specifically, our experiments are separated into five parts: (1) comparative experiment in the RAP and RAPv2 datasets, (2) transfer learning of the view predictor in the PA-100K dataset, (3) complexity analysis of the modified Inception-v4 backbone in the RAP dataset, (4) ablation study in the RAP and RAPv2 dataset, (5) pedestrian attribute recognition experiment in real surveillance scenarios.

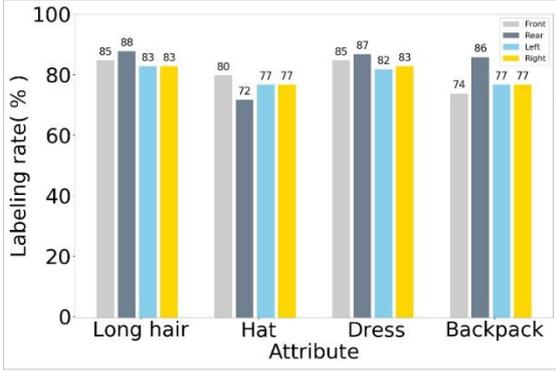

Fig. 6  Labeling rates of some typical attributes from different views in the RAPv2 dataset[5]. Typical attributes are *long hair*, *hat*, *dress*, and *backpack*.

## 4.1 Datasets

**The RAP** dataset[12] is collected from real indoor surveillance scenarios, and 26 cameras are selected to acquire all 41585 samples. Each image of this dataset is annotated with 72 fine-grained attributes and extra contextual factors, including viewpoints, occlusion, and body parts.

**The RAPv2** dataset[5] comes from a realistic surveillance scenario in a shopping mall, and all 84928 images that contain 2589 person identities, are captured by 25 cameras. The viewpoint attribute is also contained in the RAPv2 dataset.

**The PA-100K** dataset[13] has become known as the largest dataset in the pedestrian attribute recognition domain, with 100000 images in total captured from 598 real outdoor surveillance cameras. According to a ratio of 8:1:1, the PA-100K dataset is randomly split into training, validation, and testing sets, and each image is labeled with 26 attributes.

The RAP and RAPv2 datasets have peculiar viewpoint labels, and the labeling rates of the same attribute are different from different views. As shown in Fig. 6, the labeling rate of the backpack attribute in the rear view is significantly higher than that in the front view in the RAPv2 dataset. Therefore, utilizing the viewpoint labels of the RAP and RAPv2 datasets, we construct the comparative experiment to verify the performance of our entire model. The PA-100K dataset lacks viewpoint labels but has a large abundance of samples, so we realize transfer learning in the PA-100K dataset to verify the reliability of our view prediction and formed view-attributes.

## 4.2 Evaluation Metrics

Two types of evaluation metrics are adopted in our experiments. (1) Label-based: the mA[48] criterion can be formulated as:

$$mA = \frac{1}{2N}\sum_{i=1}^{M}(\frac{TP_i}{P_i}+\frac{TN_i}{N_i}) \quad (7)$$

where $N, M$ are the numbers of examples and attributes, respectively. $TP_i$ and $TN_i$ are the numbers of correctly predicted positive and negative samples of the $i$-th attribute, respectively. $P_i$ and $N_i$ are the numbers of positive and negative samples, respectively. (2) Instance-based: the criteria of accuracy, precision, recall, and F1 score[12] are adopted, which are defined as follows:

$$Accuracy = (\frac{TP_i+TN_i}{P_i+N_i}) \quad (8)$$

$$Precision = (\frac{TP_i}{TP_i+FP_i}) \quad (9)$$

$$Recall = (\frac{TP_i}{TP_i+FN_i}) \quad (10)$$

$$F_1 = (\frac{2\times Precision\times Recall}{Precision+Recall}) \quad (11)$$

where $FP_i$ and $FN_i$ are the numbers of incorrectly predicted positive and negative samples of the $i$-th attribute, respectively.

## 4.3 Loss Function

Two types of cost functions are adopted in the training process of the entire method. The first relates to view prediction, and the second relates to attribute recognition. For the first cost function, the following negative log-likelihood is used.

$$Loss_{vp} = -\frac{1}{4}\frac{1}{L}\sum_{l=1}^{L}\sum_{k=1}^{K}\sum_{t=1}^{T=4} V_{k,l}^{t} \log \widehat{V_{k,l}^{t}} \quad (12)$$

where in Eq. 12, $L$ and $K$ denote the numbers of images and attributes, respectively. $T$ is the number of views, here is equal to 4. $V_{k,l}^{t}$ indicates the ground truth label of $i$-th view of $k$-th attribute in $l$-th image, and $\widehat{V_{k,l}^{t}}$ is the assumed prediction of the view information.

The following weighted cross-entropy loss[31] is utilized as the loss for attribute recognition.

$$Loss_a = -\frac{1}{N}\sum_{i=1}^{N}\sum_{j=1}^{M} w_j(y_{i,j}\log(\sigma(\widehat{y_{i,j}}))$$

$$+(1-y_{i,j})\log 1-\sigma(\widehat{y_{i,j}}))) \quad (13)$$

$$w_j = \begin{cases} e^{1-r_j} & y_{i,j}=1 \\ e^{r_j} & y_{i,j}=0 \end{cases} \quad (14)$$

where $N$ and $M$ denote the numbers of images and attributes, respectively, $y_{i,j}$ indicates the ground truth of $j$-th attribute in $i$-th image, and $\widehat{y_{i,j}} \in \{0,1\}$ is the predicted attribute result. $w_j$ in Eq. 14 denotes the weight for $j$-th attribute to alleviate the imbalanced distribution problem of positive and negative samples[2], and $r_j$ is the proportion of positive samples of $j$-th attribute.

The final loss function is a weighted combination of these two loss functions.

$$Loss = \alpha Loss_{vp} + \beta Loss_a \quad (15)$$

where $\alpha$ and $\beta$ are hyper-parameters to balance the two different losses.

## 4.4 Implementation Details

In our experiments, the average pooling layer, the dropout layer, and the final softmax from the original Inception-v4[42] structure are eliminated, and a BN layer is added to the final attribute recognition units to balance the change. Without other tricks, only the random crop strategy is employed to augment data for avoiding data unbalanced and over-fitting. During the training process, images are resized and normalized into 256×192, and the SGD optimizer is utilized with a batch size of 64, a momentum of 0.9, and a weight decay of $5\times10^{-5}$. Our model is trained on four NVIDIA 2080Ti GPUs based on the Pytorch environment. When handling shallow view features, the learning rate is set to 0.1, while in the deeper training, the learning rate equals 0.01.

Table 1 Comparative results of our proposed method and some state-of-the-art works on the RAP dataset[12]. All methods are listed and classified into five categories in detail. The **best** results are highlighted in bold, and the second-best results are underlined.

| Dataset | RAP | | | | | |
|---|---|---|---|---|---|---|
| Method \ Metrics | mA | Accu | Prec | Recall | F1 | Category |
| ACN[1] | 69.66 | 62.61 | **80.12** | 72.26 | 75.98 | Global-based: Multi-attribute joint prediction |
| DeepMAR[31] | 73.79 | 62.02 | 74.92 | 76.21 | 75.56 | |
| PGDM[2] | 74.31 | 64.57 | 78.86 | 75.90 | 77.35 | Part-based: Pose estimation as the auxiliary supervision |
| DTM+AWK[32] | 82.04 | 67.42 | 75.87 | 84.16 | 79.80 | |
| JRL[33] | 74.74 | - | 75.08 | 74.96 | 74.62 | Relation-based: Joint recurrent learning of attribute context and correlation by LSTM module |
| GRL[34] | 81.20 | - | 77.70 | 80.90 | 79.29 | |
| HP-Net[13] | 76.12 | 65.39 | 77.33 | 78.79 | 78.05 | Attention-based: Multi-scale attention mechanism |
| Da-HAR[8] | **84.28** | 59.84 | 66.50 | 84.13 | 74.28 | Attention mask prediction |
| VeSPA[49] | 77.70 | 67.35 | 79.51 | 79.67 | 79.59 | Attribute-based: Visual prediction |
| ALM[35] | 81.87 | **68.17** | 74.71 | **86.48** | 80.16 | Attribute related-region localization |
| **Our VALA** | 78.33 | 67.48 | 79.81 | 80.84 | **80.32** | View predictor + Attention mechanism + Attribute related-region |

### 4.5 Comparative Experiment

We compare our VALA approach on the RAP dataset[12] with a great number of state-of-the-art pedestrian attribute recognition methods, e.g. ACN[1], DeepMAR[31], VeSPA[49], HP-Net[13], JRL[33], GRL[34], PGDM[2], ALM[35], Da-HAR[8], and DTM+AWK[32]. These PAR algorithms place emphasis on different aspects, including global-based, part-based, attention-based, relation-based, and attribute-based. The comparative results of our VALA and other methods on the RAP dataset are displayed in Table. 1.

The results show that our proposed method achieves competitive performances under both label-based and instance-based metrics on the RAP dataset. As for the F1 score, VALA surpasses all state-of-the-art methods with 80.32%. Though in terms of accuracy and precision, the proposed method wins the second-best results, its scores lag only 0.69% and 0.31% behind the best methods ALM[35] and ACN[1], respectively. Compared with previous methods relying on multi-attribute joint prediction, our proposed method can reach a significant improvement due to the consideration of specific attribute concerns. Our method performs a higher mA matric, about 4.02%, than the part-based method PGDM[2] has, which demonstrates the effectiveness of taking view features as the prior knowledge. Meanwhile, we find that our VALA achieves better gains than VeSPA[49], which also belongs to the view prediction method and classifies the training dataset according to different views. The main reason is that our regional attention is utilized to localize specific attributes from different views rather than distinguish different view images to find attributes manually. Better performances of our VALA can be seen from the comparison between these two methods on the PA-100K dataset. Moreover, more recent methods have begun to focus on the attention mechanism or attribute-specific localization. However, our proposed method still gets a comparable result for the capability to trace more precise attribute-corresponding regions by regional attention branch.

Table 2 Comparative results of our proposed method and some recent works on the RAPv2 dataset[5]. Results* of DIAA[6], VAC[50], and ALM[35] are collected from the reimplementation of Jia et al. [51]. Our experiments adopt the same setting as these works for a fair comparison, and the **best** results are highlighted in bold, while the second-best results are underlined.

| Dataset | RAPv2 | | | | |
|---|---|---|---|---|---|
| Method \ Metrics | mA | Accu | Prec | Recall | F1 |
| DIAA[6]* | 77.87 | 67.19 | 79.03 | 79.79 | 79.04 |
| VAC[50]* | 76.74 | **67.52** | 80.42 | 78.78 | 79.24 |
| ALM[35]* | 78.21 | 66.98 | 78.25 | 80.43 | 78.93 |
| Strong Baseline[51] | 77.34 | 66.12 | **81.99** | 75.62 | 78.21 |
| **Our VALA** | **78.30** | 67.00 | 79.10 | **81.03** | **80.05** |

The comparative experiment is also conducted on the RAPv2 dataset[5]. Since the RAPv2 dataset is not as universal as the RAP dataset[12], we only compare our method with some recent works, including DIAA[6], VAC[50], ALM[35], and Strong Baseline[51]. Results of these methods are collected from the reimplementation of Jia et al. [51], and our experiments adopt the same setting as these works for a fair comparison. As shown in Table. 2, our VALA obtains the best result in the mA metric, the recall metric, and the F1 score, which illustrates the comparability and potential of our method in recent pedestrian attribute recognition.

### 4.6 Transfer Learning Analysis

Owing to the unique viewpoint annotations of the RAP[12] and RAPv2[5] datasets, we use the viewpoint labels as the ground truth to train our view predictor. To verify the applicability and transferable capability of view prediction and the benefit of view-attribute supervision, we fix the para-meters of the view prediction branch trained on the RAP dataset[12], and retrain the attribute recognition part to implement transfer learning on the PA-100K dataset[13]. Moreover, we complement LG-Net[4] and MT-CAS[52] as comparative methods.

Table 3  Transfer learning results of our view predictor on the PA-100K dataset[13] with the bold **best** result and underline second-best result. The results are obtained by fixing the trained parameters of the view prediction branch and retraining our attribute inference part.

| Dataset | PA-100K | | | | |
|---|---|---|---|---|---|
| Method \ Metrics | mA | Accu | Prec | Recall | F1 |
| DeepMAR[31] | 72.70 | 70.39 | 82.24 | 80.42 | 81.32 |
| HP-Net[13] | 74.21 | 72.19 | 82.97 | 82.09 | 82.53 |
| VeSPA[48] | 76.32 | 73.00 | 84.99 | 81.49 | 83.20 |
| PGDM[2] | 74.95 | 73.08 | 84.36 | 82.24 | 83.29 |
| LG-Net[4] | 76.96 | 75.55 | 86.99 | 83.17 | 85.04 |
| ALM[35] | 80.68 | 77.08 | 84.21 | 88.84 | 86.46 |
| DTM+AWK[32] | **81.63** | 77.57 | 84.27 | **89.02** | 86.58 |
| MT-CAS[52] | 77.20 | 78.09 | **88.46** | 84.86 | 86.62 |
| **Our VALA** | 80.08 | **78.14** | 87.60 | 86.73 | **87.16** |

Table 4  Complexity results of the modified inceptionv4 backbone on the RAP dataset[12]. The parameters and F1 scores are produced by the whole models.

| Method | Backbone | Params | F1 |
|---|---|---|---|
| **Our VALA** | ResNet50 | 29.4M | 78.51 |
| | GoogleNet | 16.3M | 76.62 |
| | BN-Inception | 17.2M | 77.35 |
| | Inception-v3 | 50.7M | 78.93 |
| | modified Inception-v4 | 46.1M | 80.32 |
| | modified Inception-ResNet-v2 | 46.1M | 80.48 |

As shown in Table. 3, the proposed method achieves better performances on the PA-100K dataset compared with existing state-of-the-art methods. Our proposed method outperforms all previous methods in terms of accuracy and F1 score, improving 0.05% and 0.54% upon the second-best method MT-CAS[52] respectively. Meanwhile, our VALA also wins the second-best result with an 87.60% precision criterion. And VALA has a comparable mA matric, which is not much worse than the method[32, 35] recently presented. These better performances benefit from our handling of the influence of viewpoint change on the attribute recognition process by utilizing view information to focus on specific attributes and fixing the corresponding view for a specific attribute. Notably, VALA manifests a more competitive performance compared with the single visual model[49] by a significant margin. This consequence attributes to the benefit of our regional attention, which embeds inter-channel dependencies and preserves spatial attribute information to localize attribute-corresponding regions from different views. The improvement of transfer learning results on the PA-100K dataset illustrates the strong transferable capability of the view prediction branch to deal with the view change problem and the stable ability of composed view-attributes to contact shallow with deep features to guide the process of deep feature extraction.

### 4.7  Complexity Analysis

Inception-v4 is suitable to become the main feature extraction network, since Inception-v4 is sufficiently deep and possesses different convolutional kernels to capture both shallow global features and deep local features. However, the enormous parameters produced by the Inception-v4 structure and other branches are inevitable. When sophisticated parameters are deployed on devices, the speed of the training process may drop, and the time of the inference process may increase.

Therefore, it is necessary to explore the influence of the modified Inception-v4 baseline complexity on the applicability of the overall model. Keeping other settings unchanged, we only replace the Inception-v4 network with a series of other inception backbones[42-45] and ResNet50[11]. We place our view prediction branch and regional attention branch at suitable positions in other backbones as much as possible to reduce the impact of different network positions on the effect of model branches. The parameters and F1 scores in Table. 4 are produced by the whole models.

As depicted in Table. 4, we find that the proposed VALA module has a certain number of parameters that are mainly caused by the Inception-v4 baseline. Nevertheless, the modification of removing the average pooling, dropout, and the final fully-connected layer decreases the number of parameters to some extent without performance reduction. Moreover, the view prediction branch and the regional attention branch only account for few parameters and increase slightly the training time, and the generated F1 score is 80.32%. When retraining our VALA on ResNet50[11], the parameters decline, but the performance has a serious drop to 78.51% in the F1 metric, which is 1.81% lower than our model with the Inception-v4 produces. Similarly, the F1 scores are as low as 76.62% and 77.35% when the backbone of our model is replaced with GoogleNet[43] and BN-Inception[44], respectively. The main reason for this phenomenon is that the above three backbones are not sufficiently deep for view features to contact deep feature information and for the regional attention to conduct deep localization. However, Inception-v3[45] has a large number of parameters that are even more than Inception-v4 has and is inclined to be deeper. However, the performance that the model with Inception-v3 produces is worse than that of our method because Inception-v4 has a more uniform simplified architecture and more different sizes of convolutional kernels for extracting different features.

Residual connections are apt to aggregate shallow networks and make the information flow between network layers, which benefits for conducting view prediction and helping view-attributes contact shallow with deep features. We replace the Inception-v4 backbone with Inception-ResNet-v2[42] to reap the benefit of residual connections and retain the advantage of the inception network concurrently. We retrain the model on the Inception-ResNet-v2 baseline to get a slightly higher F1 score and quicker training speed. Although the model with Inception-ResNet-v2 gains better performance and computational efficiency, our method with the modified Inception-v4 network is still comparable for having the same quantity in parameters and a closely similar effect without residual connections.

### 4.8 Ablation Study

The ablative experiment is set to justify the contribution of each block in our method, and every component is append-ed gradually. As shown in Table. 5, starting with the single Inception-v4 baseline, we construct several variants to compare our method.

**The capability of prior view information:** The view feature supervision is added first based on the baseline, and an impressive promotion is found, which demonstrates the capability of the view features to construct the relevance of shallow and deep layers and supervise the network to focus on specific attributes. Moreover, the operation that makes specific attributes associated with their related views helps the view-attributes be identified easier. Then, three view weights (namely, abandon the right view) are added to the final attribute recognition units. An obvious improvement is gained, which illustrates the feasibility of considering view information as prior knowledge to guide the final attribute recognition. Then the abandoned right view is set back for the purpose of supplying full views (namely, the same view setting as our VALA). Since an input image is not flipped, attributes from the right view can be regarded as the mirror flipping for attributes from the left view to augment data. The four views module still has comparable performance considering the view completeness, though relatively few improvements are yielded by the additional right view.

**The capability of the regional attention:** On the basis of acquiring visual clues as the auxiliary supervision, we create regional attention to further assist attribute recognition by acquiring attribute location information and localizing precise attribute-related regions. To evaluate the capability of our regional attention to obtain attribute-related areas for view-attributes, we replace the regional attention with SE-Net[9] and CBAM[10] as comparisons. When SE-Net and CBAM capture view feature maps from the stacking Inception-C blocks, their effects of localizing regions are reasonably equivalent, but far from the effect performed by our regional attention. This fact verifies that embedding inter-channel dependencies and preserving spatial information in one process truly have an important impact on the aspect of attribute-corresponding region localization, which is crucial for a more precise attribute recognition process. Furthermore, the contribution of internal branches of the regional attention is also explored, and we construct two variants for comparison. Compared to our VALA, the F1 score drops to 78.06%, with a 1.24% margin when the width branch is removed, while the F1 score of that without the height branch is lower than our method with a 0.54% margin. The reason for this phenomenon is that capturing the spatial information of a specific view attribute plays a more significant role in attribute-related region localization.

Table 5 Ablation study results on the RAPv2 dataset[5]. Here, VF denotes embedding view feature supervision to network layers. VP denotes adding view weights to final attribute recognition units, while VFB+VPB is the operation that uses features from the Inception-B blocks to generate view features and view prediction. RA is our regional attention, and RAB is the operation that uses the features from the Inception-B blocks as the input of the regional attention. RAH is the regional attention without the width branch. Similarly, RAW is the regional attention without the height branch.

| Model \ Metric | mA | Accu | Prec | Recall | F1 |
|---|---|---|---|---|---|
| Baseline | 74.32 | 62.97 | 77.02 | 76.58 | 76.80 |
| Baseline+VF | 75.92 | 64.33 | 77.10 | 77.92 | 77.51 |
| Baseline+VF+3 VP | 75.98 | 64.52 | 77.21 | 78.03 | 77.62 |
| Baseline+VF+4 VP | 76.20 | 64.74 | 77.21 | 78.34 | 77.77 |
| Baseline+RA | 76.32 | 65.30 | 77.93 | 78.50 | 78.21 |
| Baseline+VF+4 VP+SE[9] | 76.31 | 65.34 | 78.01 | 78.51 | 78.26 |
| Baseline+VF+4 VP+CBAM[10] | 76.39 | 65.38 | 78.12 | 78.52 | 78.32 |
| Baseline+VF+4 VP+RAH | 75.90 | 64.79 | 77.72 | 78.40 | 78.06 |
| Baseline+VF+4 VP+RAW | 76.88 | 65.83 | 78.19 | 79.34 | 78.76 |
| Baseline+VFB+4 VPB+RA | 76.98 | 65.90 | 78.22 | 78.91 | 78.56 |
| Baseline+VF+4 VP+RAB | 76.51 | 65.67 | 77.83 | 78.80 | 78.31 |
| **VALA(Ours)** | **77.38** | **66.23** | **78.80** | **79.80** | **79.30** |

**The contribution of two branches in the whole method:** In this experiment, we evaluate the contribution of the two branches, and each branch is appended gradually. Compared with the model without viewpoint component (Baseline+RA), the result of our VALA is explicitly improved in all evaluation criteria with 1.06% in mA, 0.93% in accuracy, 0.87% in precision, 1.30% in recall, and 1.09% in F1 score. The greater result demonstrates the feasibility of utilizing view information to handle the viewpoint change problem in the attribute recognition process. When regional attention is removed (Baseline+VF+4VP), a more severe drop has happened even though the other setting is kept unchanged, further explaining that visual attribute localization is significant for recognizing a specific attribute. Regarding the above two variants with either view information or regional attention, we find that the performance improvement produced by regional attention is higher than that generated by the view prediction. The remarkable evaluation indicates that it is still necessary to accurately localize the view-attribute after capturing it for a better recognition effect. Visualization of some attribute attentive areas in Fig. 7 shows the contribution of two modules specifically.

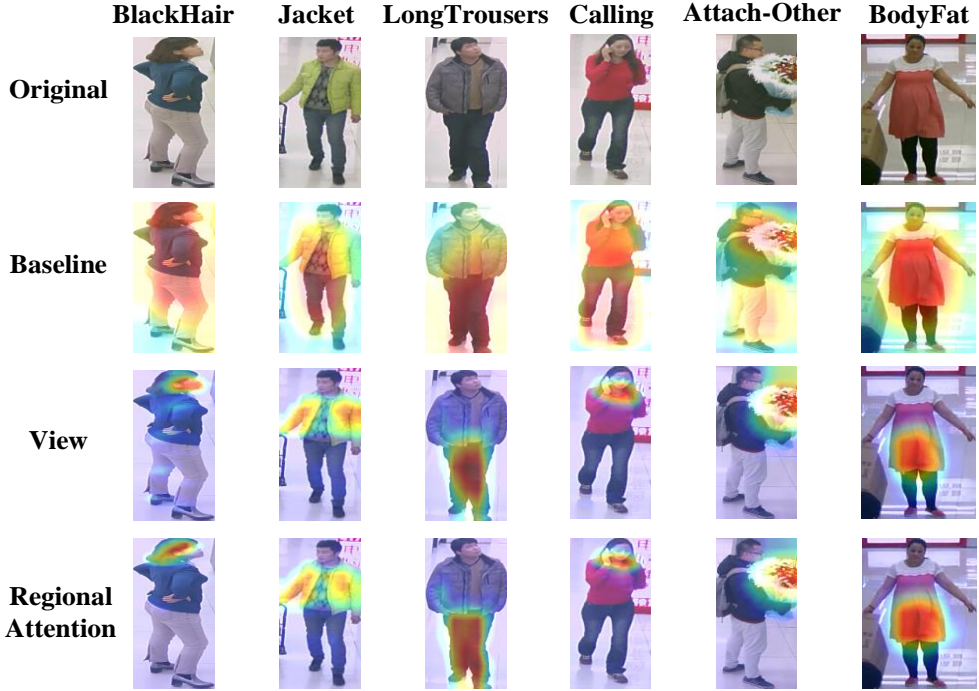

Fig. 7 Attention heatmaps generated by the single baseline, view supervision, and regional attention. The example images are from the RAPv2 dataset[5], and the attributes to be located are *Black Hair*, *Jacket*, *Long Trousers*, *Calling*, *Attach-Other*, *Body Fat* in each column.

**The layers to export features for two branches:** In our original implementation, we get shallow feature maps for the view prediction from the output of stacking Inception-A blocks, while deep feature maps are exported from stacking Inception-C blocks. According to the empirical thought, the Inception-A and Inception-C layers in the entire inception network are thought to be either shallow or deep enough for view extraction and attention localization. However, it is still necessary to explore the feature maps from which layer can meet the more satisfying demand. Therefore, we rearrange the position of the view prediction branch to the end of the Inception-B blocks and keep the attention branch position unchanged. Compared with our original setting, a dropping result indicates that feature maps in Inception-A are more holistic and appropriate for extracting global view information. Referring to the first experimental result, the position of the view prediction is retained, but the regional attention is assigned to the end of the Inception-B blocks. The performance of our VALA still surpasses that of the variant, which confirms that the encoding channel and spatial attribute information of feature maps in relatively deeper layers can reach a better attribute region localization.

To sum up, ablation studies show that it is important to consider the view information as an effective prior knowledge, which is conducive to improving of the pedestrian attribute recognition model performance. Meanwhile, the capability of the regional attention to localize precise attribute-related regions increases the fine-grained and overall performance of the attribute recognition.

### 4.9 Pedestrian attribute recognition in Surveillance

In order to verify the actual recognition effect of our VALA model, we select pedestrian images from real video surveillance scenarios to test our model. Our model is pretrained on the RAP[12] dataset, and some of the representative results are shown in Fig. 8. The results illustrate that even though no positioning information is included in the pedestrian images, the VALA model is also able to accomplish the identification of the same attribute from different views. For example, the view of the glasses attribute is side-view in the picture (a) and front-view in (c), but both are successfully identified. Moreover, some local attributes such as *Black Hair* have also been recognized, which displays the accuracy for fine-grained classification by our model.

The VALA model adopts a multi-task learning framework as a whole, improving the identification performance of all attributes. It forecasts the pedestrian perspective through the view prediction, and then practices specific view-attribute recognition units. Compared with the localization of body parts, the view prediction based on global view features is easier to train, and the calculation cost is lower. Moreover, attribute classification units leverage the channel and spatial attention mechanism to focus on the most representative

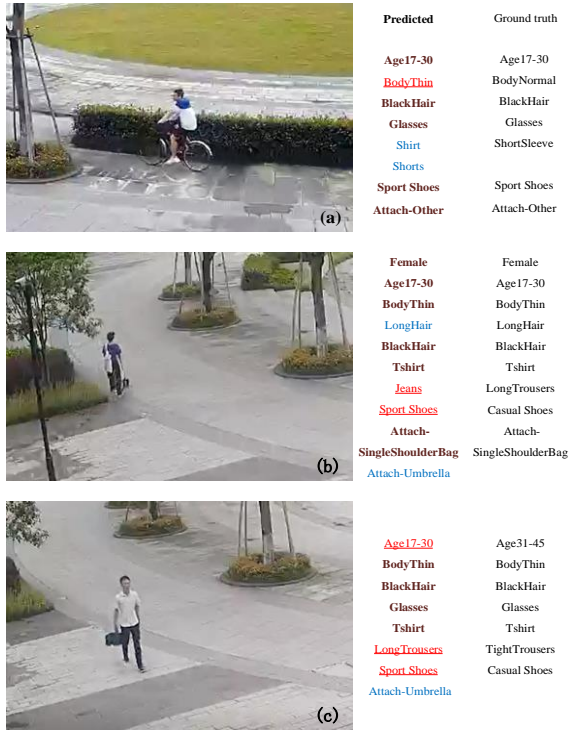

Fig. 8 Attribute recognition results of pedestrian images in real video surveillance scenarios. **Correct** predictions are marked in bold font and brown, missed attributes are in blue, incorrect predictions are in red underline format, and ground truths are in black.

view-attribute positions, so that the attribute recognition performance is effectively improved.

## 5  Conclusion

We propose an attention-based pedestrian attribute recognition model, which can predict view information to compose view-attributes with the view predictor and localize specific regions of view-attributes by the attention mechanism. The view prediction branch is trained using the unique view tags of the RAP and RAPv2 datasets and the shallow convolutional features of the inception network. To construct view-specific attribute recognition units, we introduce a channel and spatial attention mechanism to enhance feature discrimination and localize specific view-attributes to ameliorate the recognition performance. In the comparative experiment on the RAP and RAPv2 datasets, multiple indicators of the VALA model reach comparable results, which confirms the overall better performance of our model. Furthermore, on the PA-100K dataset, the applicability and stability of the view prediction and the view-attribute supervision are verified by evaluating transfer learning. In future research, we will pay more attention to the deployment possibilities of the pedestrian attribute recognition network model based on the framework of this article on mobile devices to achieve greater practical application value.

## Acknowledgements

This work was supported by The National Key R&D Program of China (2018YFB1308000) and The Natural Science Foundation of Zhejiang province (LY21F030018).